\documentclass{article}

\usepackage{times}
\usepackage{graphicx} 
\usepackage{subcaption}

\usepackage{natbib}

\usepackage{algorithm}
\usepackage{algorithmic}

\usepackage{hyperref}

\usepackage[accepted]{icml2016} 
\usepackage{amsmath,amsfonts,amssymb}
\usepackage{dsfont}
\usepackage{booktabs}
\usepackage{relsize}
\usepackage{lmodern}
\usepackage{slantsc}
\usepackage{siunitx}
\newcommand{\xsj}{\mathbf{\hat{x}}_j}
\newcommand{\xuj}{\mathbf{x}_j}
\newcommand{\xsi}{\mathbf{\hat{x}}_i}
\newcommand{\xui}{\mathbf{x}_i}
\newcommand{\ysi}{\hat{y}_i}

\newcommand{\yui}{y_i}
\newcommand{\sw}{s_{ \textsc{\relsize{-2}{\textsl{W}}} }}

\icmltitlerunning{Unsupervised Transductive Domain Adaptation}
\begin{document} 

\twocolumn[
\icmltitle{Unsupervised Transductive Domain Adaptation}

\icmlauthor{Ozan Sener}{ozansener@cs.stanford.edu}
\icmladdress{Stanford University and Cornell University}
\icmlauthor{Hyun Oh Song}{hsong@cs.stanford.edu}
\icmladdress{Stanford University} 

\icmlauthor{Ashutosh Saxena}{ashutosh@brainoft.com}
\icmladdress{Brain of Things} 

\icmlauthor{Silvio Savarese}{ssilvio@stanford.edu}
\icmladdress{Stanford University } 

\icmlkeywords{domain adaptation, transductive learning, metric learning, deep learning}

\vskip 0.3in
]

\begin{abstract} 
Supervised learning with large scale labeled datasets and deep layered models has made a paradigm shift in diverse areas in learning and recognition. However, this approach still suffers generalization issues under the presence of a domain shift between the training and the test data distribution. In this regard, unsupervised domain adaptation algorithms have been proposed to directly address the domain shift problem. In this paper, we approach the problem from a transductive perspective. We incorporate the domain shift and the transductive target inference into our framework by jointly solving for an asymmetric similarity metric and the optimal transductive target label assignment. We also show that our model can easily be extended for deep feature learning in order to learn features which are discriminative in the target domain. Our experiments show that the proposed method significantly outperforms state-of-the-art algorithms in both object recognition and digit classification experiments by a large margin.
\end{abstract} 

\section{Introduction}
\label{intro}
Recently, deep convolutional neural networks \cite{alexnet, vggnet, googlenet} have propelled unprecedented advances in artificial intelligence including object recognition, speech recognition, and image captioning. One of the major drawbacks of the method is that the network requires a lot of labelled training data to fit millions of parameters in the complex network model. However, creating such datasets with complete annotations is not only tedious and error prone, but also extremely costly. In this regard, the research community has proposed different mechanisms such as semi-supervised learning \cite{semisup1,semisup2,semisup3}, transfer learning \cite{transfer1, transfer2}, weakly labelled learning, and domain adaptation. Among these approaches, domain adaptation is one of the most appealing techniques when a fully annotated dataset (e.g. ImageNet \cite{ImageNet}, Sports1M \cite{sports1m}) is available as a reference. 

Formally, the goal of unsupervised domain adaptation is: given a fully labeled source dataset and an unlabeled target dataset, to learn a model which can generalize to the target domain while taking the domain shift across the datasets into account. The majority of the literature \cite{gong12, baochen15, fernando13, baochen16, tommasi13} in unsupervised domain adaptation formulates a learning problem where the task is to find a transformation matrix to align the labelled source data distribution to the unlabelled target data distribution. Although these approaches show promising results, they do not take the actual target inference procedure into the learning algorithm. We solve this problem by incorporating the unknown target labels into the training procedure.

Concretely, we formulate a unified framework where the domain transformation parameter and the target labels are jointly optimized in two alternating stages. In the transduction stage, given a fixed domain transform parameter, we jointly infer all target labels by solving a discrete multi-label energy minimization problem. In the adaptation stage, given a fixed target label assignment, we seek to find the optimal asymmetric metric  between the source and the target data. The advantage of our method is that we can learn a domain transformation parameter which is aware of the subsequent transductive inference procedure. 

Following the standard evaluation protocol in the domain adaptation community, we evaluate our method on the digit classification task using MNIST \cite{mnist} and SVHN\cite{svhn} as well as the object recognition task using the Office \cite{office} dataset, and demonstrate state of the art performance in comparison to all existing unsupervised domain adaptation methods.

\section{Related Work} 

This paper is closely related to two active research areas: (1) Unsupervised domain adaptation, and (2) Transductive learning.

\textbf{Unsupervised domain adaptation}: \cite{gong12, baochen15, fernando13, baochen16} proposed subspace alignment based approaches to unsupervised domain adaptation where the task is to learn a joint transformation and projection where the difference between the source and the target covariance is minimized. However, these methods learn the transform matrices on the whole source and target dataset without utilizing the source labels. 

\cite{tommasi13} utilizes local max margin metric learning objective \cite{lmnn} to first assign the target labels with the nearest neighbor scheme and then learn a distance metric to enforce that the negative pairwise distances are larger than the positive pairwise distances. However, this method learns a symmetric distance matrix shared by both the source and the target domains so the method is susceptible to the discrepancies between the source and the target distributions. Recently, \cite{ganin15, tzeng14} proposed a deep learning based method to learn domain invariant features by providing the reversed gradient signal from the binary domain classifiers. Although this method perform better than aforementioned approaches, their accuracy is limited since domain invariance does not necessarily imply discriminative features in the target domain. 

\textbf{Transductive learning}: In the transductive learning \cite{transduction}, the model has access to unlabelled test samples during training. Recently, \cite{coclassification} tackled a classification problem where predictions are made jointly across all test examples in a transductive \cite{transduction} setting. The method essentially enforces the notion that the true labels vary smoothly with respect to the input data. We extend this notion to infer the labels of unsupervised target data in a k-NN graph. 

To summarize, our main contribution is to formulate a joint optimization framework where we alternate between inferring target labels via discrete energy minimization (\textit{transduction}) and learning an asymmetric transformation (\textit{adaptation}) between source and target examples. Our experiments on digit classification using MNIST \cite{mnist} and SVHN\cite{svhn} as well as the object recognition experiments on Office \cite{office} datasets show state of the art results outperforming all existing methods by a substantial margin.

\section{Method} 
We address the problem of unsupervised transductive domain adaptation by jointly solving for the label assignment of unsupervised target domain as well as the shift between the domains. We first define our model in Section~\ref{prob:def} and explain the two sub-problems of transduction and adaptation. We further explain the details of transduction in Section~\ref{label} and the details of adaptation in Section~\ref{metric}.

\subsection{Problem Definition}
\label{prob:def}
In the unsupervised domain adaptation problem, one of the domains (source) is fully supervised $\{\xsi, \ysi \}_{i \in [N^s]}$ with $N^s$ data points $\xsi$ and corresponding labels $\ysi$ from a discrete set $\ysi \in \{1,\ldots, k \}$.  The other domain (target), on the other hand is unsupervised and has $N^u$ data points $\{\xui \}_{i \in [N^u]}$. 

We further assume that both domains have different distributions $\xsi \sim p_s$ and $\xui \sim p_t$ defined on the same space as $\xsi,\xui \in \mathcal{X}$ and there exists a feature function \mbox{$\Phi:\mathcal{X}\rightarrow \mathcal{R}^d$} which is applicable to both. We further study the case where the feature function is parametric with a parameter $\mathbf{\theta}$ defined as \mbox{$\Phi_\mathbf{\theta}:\mathcal{X}\rightarrow \mathcal{R}^d$}, and we develop a method to learn the parameters.

Our model has two main components, transduction and adaptation. The transduction is the sub-problem of labelling unsupervised data points and the adaptation is solving for the domain shift. 

For adaptation, we explicitly model the domain shift in the form of an asymmetric similarity as
\begin{equation}
\sw(\xsi,\xuj) = \Phi(\xsi)^\intercal \mathbf{W} \Phi(\xuj)
\end{equation}
such that it is high if two data points, $\xsi$ from the source and $\xuj$ from the target, are from the same class.

We further model our transduction in the form of a nearest neighbor and we follow the triplet loss defined in \cite{lmnn} in order to solve adaptation by taking the nearest neighbor inference into learning. While the original triplet loss \cite{lmnn} enforces a margin $\alpha$ between the similarity of any point to its nearest neighbor from the same class and the nearest neighbor from other classes, we extend this construction to the unsupervised domain adaptation by enforcing a similar margin. For each source point, we enforce a margin between its similarity with the nearest neighbor from the target having the same label and having a different label as; $ \sw(\xsi,\mathbf{x}_{i^+}) > \sw(\xsi,\mathbf{x}_{i^-}) + \alpha$ where $\mathbf{x}_{i^+}$ is the nearest target having the same class as $\xsi$ and $\mathbf{x}_{i^-}$ is the nearest target having a different class label.

Since we model our problem as transduction, we include target labels as part of the joint learning and introduce a target label consistency term as well. We enforce that similar unsupervised data points should have the same label after the transduction by penalizing label disagreements between similar images.

Our model leads to the following optimization problem, over the target labels $\yui$ and the similarity metric $\mathbf{W}$, jointly solving transduction and adaptation. 
\begin{equation}
\begin{aligned}
\min_{\mathbf{W}, y_1, \ldots y_{N^u}} &\sum_{i \in [N^s]} &&[\sw(\xsi,\mathbf{x}_{i^-}) - \sw(\xsi,\mathbf{x}_{i^+}) + \alpha]_{+}  \\
&+\lambda &&\hspace{-3mm}\sum_{i \in N^u} \sum_{j \in \mathcal{N}(\xui)}  sim(\xui, \xuj) \mathds{1}(y_i \neq y_j)\\
&s.t. \quad &&i^{+} = {\arg\max}_{j | y_j = \hat{y}_i} \sw(\mathbf{\hat{x}}_i,\mathbf{x}_{j}) \\
&\quad &&i^{-} = {\arg\max}_{j | y_j \neq \hat{y}_i} \sw(\mathbf{\hat{x}}_i,\mathbf{x}_{j}) 
\end{aligned}
\label{loss}
\end{equation}
where $\mathds{1}(a)$ is an indicator function which is $1$ if $a$ is true and $0$ otherwise. $[a]_+$ is a rectifier function which is equal to $\max(0, a)$, and $sim$ is any similarity function. We use cosine similarity as $sim(\xui, \xuj)= \frac{\Phi(\xui)^\intercal \Phi(\xuj)}{|\Phi(\xui)||\Phi(\xuj)|}$. 

We solve this optimization problem via alternating minimization through iterating over solving for unsupervised labels $y_i$(transduction) and learning the similarity metric $\mathbf{W}$ (adaptation). We explain these two steps in detail in the following sections.



\subsection{Transduction: Labeling Target Domain}
\label{label}
\begin{figure}[ht]
\includegraphics[width=\columnwidth]{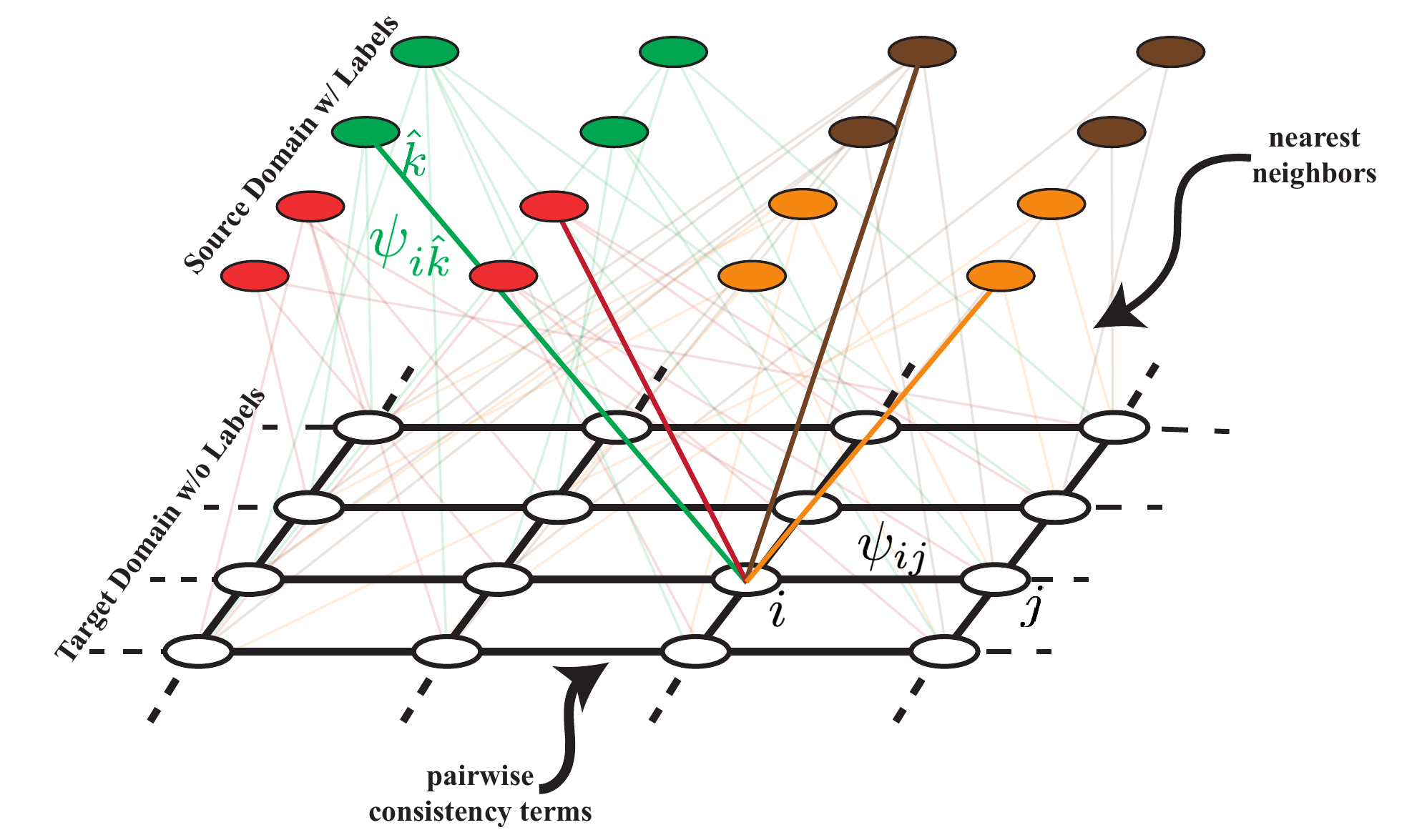}
\vspace{-6mm}
\caption{\textbf{Visualization of the Label Propogation.} We create a k-NN graph of unsupervised target points to enforce consistency with pairwise terms 
\mbox{$\psi_{ij}=sim(\mathbf{x}_i, \mathbf{x}_j) \mathds{1}(y_i \neq y_j)$} and use closest supervised source points to each class as 
\mbox{$ \psi_{i\hat{k}} \sw(\mathbf{\hat{x}}_k,\mathbf{x}_{i})$}.} 
\label{vis_label_prop}
\end{figure}
In order to label the unsupervised points, we use the nearest-neighbor rule. We simply compute the NN supervised data point for each unsupervised data point using the learned metric $\sw(\cdot,\cdot)$ and transfer the corresponding label. In the initial stages of our optimization, our tranduction needs to be accurate even with a sub-optimal similarity metric due to the iterative fashion of our algorithm. Hence, we enforce a consistent labeling via label propagation. We first formally define the NN-rule and then introduce the label propagation.

Given a similarity metric $\sw(\cdot,\cdot)$, the NN rule is:
\begin{equation}
(y_i)^{pred} = \hat{y}_{{\arg\max}_j \sw(\mathbf{x_i}, \mathbf{\hat{x}_j})}
\end{equation}

We use label propagation to enforce consistency of the predicted labels of unsupervised data points. Our label propagation is similar to existing graph transduction algorithms \cite{label_prop1,label_prop2}. In order to enforce this consistency, we create a k-nearest neighbor (k-NN) graph over the unsupervised data points such that neighbors $\mathcal{N}(\xui)$ for $\xui$ is the k-unsupervised data point having highest similarity to $\xui$ using the cosine similarity in the feature space. After the k-NN graph is created, we solve the following optimization problem for labeling unsupervised data points with label propagation:
\begin{equation}
\begin{aligned}
\min_{y_1, \ldots y_{N^u}}  &\sum_{i \in N^u} - \max_{\hat{y}_j=y_i}  \sw(\mathbf{\hat{x}}_j,\mathbf{x}_{i}) \\
&+ \lambda
\sum_{i \in N^u} \sum_{j \in \mathcal{N}(\mathbf{x}_i)} sim(\mathbf{x}_i, \mathbf{x}_j) \mathds{1}(y_i \neq y_j)
\end{aligned}
\label{robtran}
\end{equation}
This problem can approximately be solved using many existing methods such as $\alpha$-$\beta$ swapping, quadratic pseudo-boolean optimization (QPBO) and linear programming through roof-duality. We use the $\alpha$-$\beta$ swapping algorithm from \cite{kolmogrovalphabeta} since it is experimentally shown to be efficient and accurate. In order to further explain the label propagation, we visualize an example with $k=4$ and $4$-class classification problem in Figure~\ref{vis_label_prop}. 

It is also critical that this formulation requires solving high number of nearest neighbors which is computationally challenging. However, our choice of optimization method makes this computation tractable. We use stochastic gradient descent in our adaptation stage with a carefully chosen batch size, which requires us to only solve the transduction over a batch. 

\subsection{Adaptation: Learning the Metric}
\label{metric}
Given the predicted labels $y_i$ for unsupervised data points $\xui$, we need to learn an asymmetric metric in order to minimize the loss function defined in (\ref{loss}). 

The main intuition behind our formulation is to seek a metric which can label the supervised points correctly using the unsupervised points and their predicted labels. In other words, we reverse the labeling direction. At this stage we have already predicted a label for each unsupervised point; hence, we can estimate a label for each supervised point using the predicted labels. We also have ground truth labels for the supervised points and we combine them to find an asymmetric metric. In other words, the goal of the adaptation stage is:

\vspace{-3mm}
\begin{itemize}
\item predicting $\hat{y}^{pred}_j$ using $\xui, \xsj, \yui$
\item learning $\sw(\cdot,\cdot)$ by penalizing  $(\hat{y}_j)^{pred} \neq \hat{y}_j$ 
\end{itemize}

Fortunately, this can be jointly solved by minimizing the triplet loss defined with supervised data points and their closest same class and different class neighbors among the unsupervised points. Formally, we find the closest same class and different class points as;
\begin{equation}
\begin{aligned}
&i^{+} = {\arg\max}_{j | y_j = \hat{y}_i} \sw(\xsi, \xui) \\
&i^{-} = {\arg\max}_{j | y_j \neq \hat{y}_i} \sw(\xsi,\xuj) 
\label{sup_nn}
\end{aligned}
\end{equation}

We further define the loss function with a regularizer using the nearest neighbors as: $loss(\mathbf{W})=$
\begin{equation}
 \sum_{i \in [N^s]} [\sw(\mathbf{\hat{x}}_i,\mathbf{x}_{i^-}) - \sw(\mathbf{\hat{x}}_i,\mathbf{x}_{i^+}) + \alpha]_{+} + r(\mathbf{W})
\end{equation}
which is convex in terms of the $\mathbf{W}$ if the regularizer is convex; and we optimize it via stochastic gradient descent using the subgradient \
\mbox{$\frac{\partial loss (y_i, \mathbf{W})}{\partial \mathbf{W}} = \frac{\partial r ( \mathbf{W})}{\partial \mathbf{W}} + $}
\begin{equation}
\begin{aligned}
\sum_{i \in [N^s]} &\mathds{1}(\sw(\xsi,\mathbf{x}_{i^-}) - \sw(\xsi,\mathbf{x}_{i^+})>\alpha) \\
&\times \left( \Phi(\xsi)\Phi(\mathbf{x}_{i^-})^\intercal - \Phi(\xsi)\Phi(\mathbf{x}_{i^+})^\intercal  \right)  
\end{aligned}
\label{gradw}
\end{equation}
As a regularizer use the Frobenius norm of the similarity matrix as $r(\mathbf{W})=\frac{1}{2}\|\mathbf{W}\|_F^2$. We explain the details of this optimization routine in the Section~\ref{imp_det}.
\subsection{Learning Features}
In Section~\ref{label}~and~\ref{metric}, we explained our transduction with label propagation as well as the adaptation algorithm using a pre-defined feature function $\Phi$. However, the current trends in machine learning suggest that learning this feature function $\Phi$ from the data using deep neural networks is a promising direction especially for visual domains. Hence, we consider the case where $\Phi_{\mathbf{\theta}}$ is a parametrized feature function with parameter set, $\mathbf{\theta}$. A typical example is CNNs (Convolutional Neural Networks) with $\mathbf{\theta}$ as concatenation of weights and biases in the layers of CNN. We learn the feature function parameters as part of the adaptation stage with an update; $\frac{\partial loss (y_i, \mathbf{W})}{\partial \mathbf{\theta}} =$

\begin{small}
\begin{equation}
\begin{aligned}
 \sum_{i \in [N^s]} &\mathds{1}(\sw(\xsi,\mathbf{x}_{i^-}) - \sw(\xsi,\mathbf{x}_{i^+})>\alpha)  \\
 &\times \left(\frac{\partial \sw(\xsi,\mathbf{x}_{i^-}) }{\partial \mathbf{\theta}} - \frac{\partial \sw(\xsi,\mathbf{x}_{i^+}) }{\partial \mathbf{\theta}} \right)
 \label{gradt}
 \end{aligned}
\end{equation}
\end{small}
where {\small $\frac{\partial \sw(\xsi,\xuj) }{\partial \mathbf{\theta}} =\Phi(\xuj)^\intercal \mathbf{W}^\intercal \frac{\partial \phi_\mathbf{\theta}(\xsi)}{\partial \theta} + \Phi(\xsi)^\intercal \mathbf{W} \frac{\partial \phi_\mathbf{\theta}(\xuj)}{\partial \theta} $}

\begin{algorithm}[tb]
   \caption{Transduction with Domain Shift}
   \label{alg:example}
\begin{algorithmic}
   \STATE {\bfseries Input:} source $\mathbf{x}_i$, target $\mathbf{\hat{x}}_i$, $y_i$, batch size $B$
   \REPEAT
   \STATE  Sample $\{\mathbf{x^b}_{1 \cdots B}\}$, $\{\mathbf{\hat{x}}^b_{1 \cdots B}, \hat{y}^b_{1\cdots B}\}$
   \STATE Solve (\ref{robtran}) for $\{y_{1 \cdots B}\}$
   \FOR{$i=1$ {\bfseries to} $B$}
      \IF{$ \hat{y}_i \textbf{ in } y_{1 \cdots y_B} $} 
   \STATE Compute ($i^+, i^-$) using $\{y_{1 \cdots B}\}$ in (\ref{sup_nn})
   \STATE Update $\frac{\partial loss}{\partial \mathbf{\theta}}$ and  $\frac{\partial loss}{\partial \mathbf{W}} $ using (\ref{gradw},\ref{gradt})
   \ENDIF
   \ENDFOR
   \STATE $\mathbf{W} \leftarrow \mathbf{W} + \alpha \frac{\partial loss (y_i, \mathbf{W})}{\partial \mathbf{W}}$ 
   \STATE $\mathbf{\theta} \leftarrow \mathbf{\theta} + \alpha \frac{\partial loss (y_i, \mathbf{W})}{\partial \mathbf{\theta}}$
   \UNTIL CONVERGENCE or $MAX\_ITER$
\end{algorithmic}
\end{algorithm}

\section{Experimental Results}
We evaluate our algorithm on various unsupervised domain adaptation tasks while focusing on two different problems, hand-written digit classification and object recognition. For each experiment, we use three domains and evaluate all adaptation scenarios.

\begin{table*}[ht]
\vspace{-3mm}
\caption{Accuracy of our method and the state-of-the-art algorithms on Office dataset and various adaptation settings}
\label{tab:res}
\begin{sc}
\begin{center}
\begin{small}
\begin{tabular}{@{}rcccccc@{}} \toprule 
 Source & Amazon & D-SLR & Webcam & Webcam &Amazon & D-SLR \\
 Target & Webcam & Webcam & D-SLR & Amazon & D-SLR & Amazon \\
 \midrule
GFK \cite{gong2012} & $.398$ & $.791$ & $.746 $ & $.371$ & $.379$ & .379   \\
SA* \cite{fernando13} & $.450$ & $.648$ & $.699$ & $.393$ & $.388$ & $.420$ \\
DLID \cite{chopra13} & $.519$ & $.782$ & $.899$ & -&- &- \\
DDC \cite{tzeng14} & $.618$ & $.950$ & $.985$ & $.522$ & $.644$& $.521$\\
DAN \cite{wang15} & $.685$ & $.960$ & $.990$ & $.531$ & $.670$ & $.540$ \\
Backprop \cite{ganin15} & $.730$ &$\mathbf{.964}$ & $\mathbf{.992}$ & $.536$ & $.728$ & $.544$\\
\midrule
Source Only & $.642$ & $.961$ & $.978$ & $.452$ & $.668$ & $.476$ \\
Our Method & $\mathbf{.804}$ &.962 & $.989$ & $\mathbf{.625}$ & $\mathbf{.839}$ & $\mathbf{.567}$ \\
 \bottomrule
\end{tabular}
\end{small}
\end{center}
\end{sc}

\end{table*}
\begin{table}[ht]
\caption{Accuracy on the digit classification task.}
\label{tab:res2}
\begin{sc}
\begin{small}
\resizebox{\columnwidth}{!}{%
\begin{tabular}{@{}r@{\hskip 1mm}c@{\hskip 1mm}c@{\hskip 1mm}c@{\hskip 1mm}c@{}} \toprule 
Source & M-M & MNIST  & SVHN & MNIST \\
Target&  MNIST & M-M & MNIST & SVHN\\
 \midrule
SA* \cite{fernando13}& $.523$ & $.569$ & $.593$ & $.211$ \\
BP \cite{ganin15} &$.732$ & $.766$ & $.738$ & $.289$ \\
\midrule
Source Only  & $.483$ & $.522$  &.549 & $.162$  \\
Our Method & $\mathbf{.835}$ & $\mathbf{.855}$ & $\mathbf{.774}$ & $\mathbf{.323}$\\
 \bottomrule
\end{tabular}}
\end{small}
\end{sc}
\end{table}

\subsection{Dataset}
We use MNIST\cite{mnist}, Street View House Number\cite{svhn} and the artificailly generated version of MNIST -MNIST-M- \cite{ganin15} to experiment our algorithm on the digit classification task. MNIST-M is a simply a blend of the digit images of the original MNIST dataset and the color images of BSDS500\cite{bsds500} following the method explained in \cite{ganin15}. Since the dataset is not distributed directly by the authors, we generated the dataset using the same procedure and further confirmed that the performance is the same as the one reported in \cite{ganin15}. Street View House Numbers dataset is a collection of house numbers collected directly from Google street view images. Each of these three domains are quite different from each other and among many important differences, the most significant ones are MNIST being grayscale and the others being colored, and SVHN images having extra confusing digits around the centered digit of interest. Moreover, all three domains are large-scale having at least 60k examples over 10 classes. 

In addition, we use the Office\cite{office} dataset to evaluate our algorithm on the object recognition task. Office dataset includes images of the objects taken from Amazon, captured with a webcam and captured with a D-SLR. Differences between domains include the white background of Amazon images vs realistic backgrounds of webcam images, and the resolution differences. The Office dataset has fewer images, with a maximum of 2478 per domain with 31 classes. 

\begin{figure}[ht]
\begin{small}
\includegraphics[width=\columnwidth]{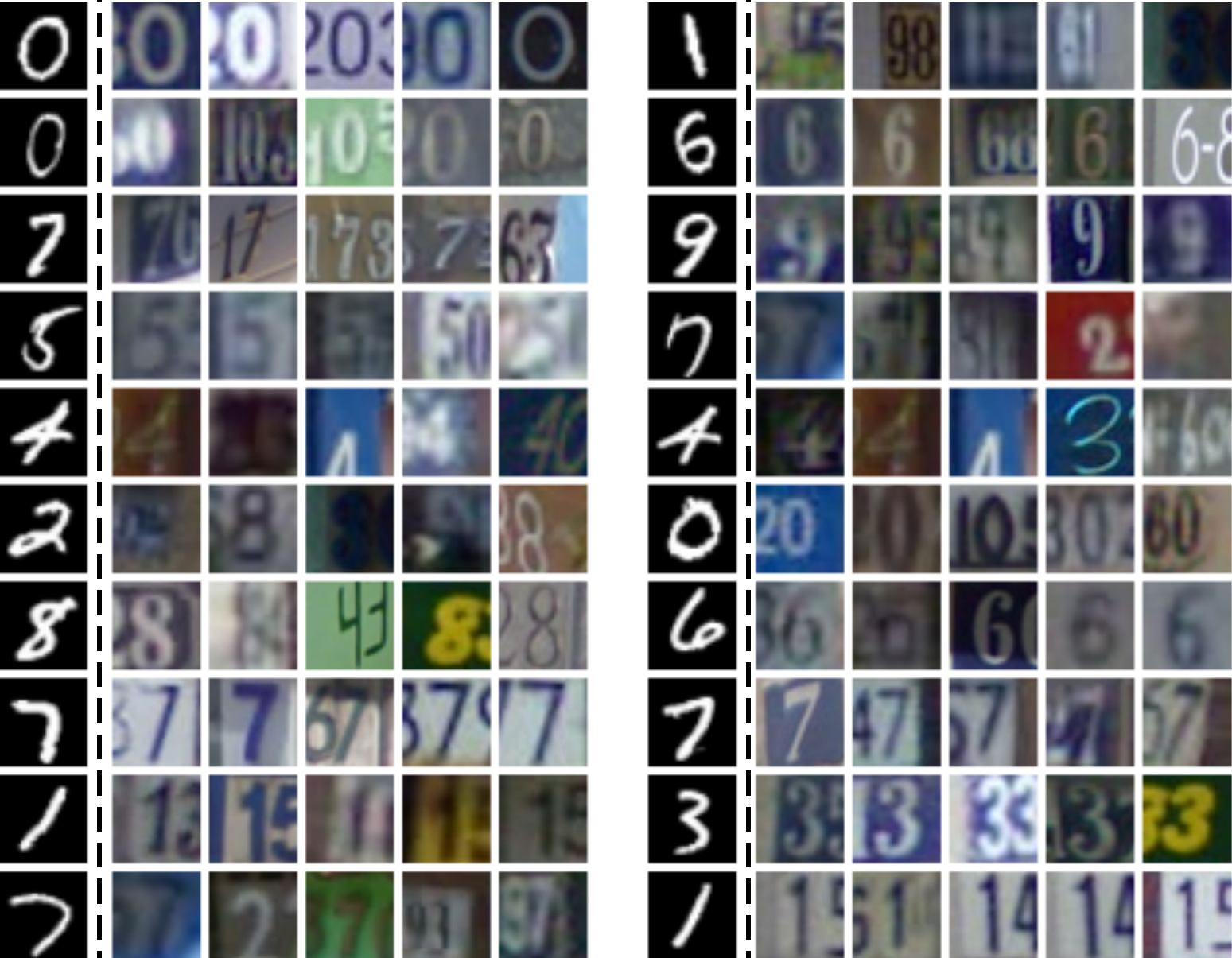}
\vspace{-5mm}
\caption{Example nearest neighbors for SVHN$\rightarrow$MNIST experiment. We show an example MNIST image and 5-NN SVHN images. Please note the large domain difference.}
\label{fig:nn}
\end{small}
\end{figure}

\subsection{Baselines}
We compare our method with a variety of methods with and without feature learning. Considering the two different lines of work, \textbf{SA*}\cite{fernando13} is the dominant state-of-the-art approach not employing any feature learning, and \textbf{Backprop(BP)}\cite{ganin15} is the dominant state-of-the-art employing feature learning. We use the available source code of \cite{ganin15} and \cite{fernando13} and following the evaluation procedure in \cite{ganin15}, we choose the hyper-parameter of \cite{fernando13} as the highest performing one among various alternatives. We also compare our method with the \textbf{source only} baseline which is a convolutional neural network trained only using the source data. This classifier is clearly different from our nearest neighbor classifier; however, we experimentally validated that CNN always outperformed the nearest neighbor based classifier. Hence, we report the highest performing source only method.

\subsection{Implementation Details}
\label{imp_det}
Although our algorithm has very few hyper-parameters and we choose most of them either using cross-validation or exhaustive grid search, our algorithm uses an existing differentiable feature function. Following the unparalleled success of convolutional neural networks (CNNs), we use CNNs as our feature functions.  In order to have a fair comparison with existing algorithms, we follow the same architecture used by \cite{ganin15} by only changing the final feature dimensionality (embedding size). We use the following architectures for domains:

\noindent \textbf{MNIST} and \textbf{SVHN:} LeNet\cite{lenet} as

\includegraphics[width=\columnwidth]{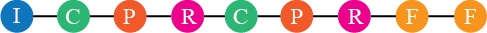}

\noindent \textbf{Office:} AlexNet\cite{alexnet} as

\includegraphics[width=\columnwidth]{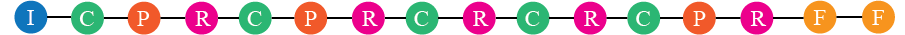}

where \textbf{C} is convolution, \textbf{P} is max-pooling, \textbf{R} is ReLU and \textbf{F} is fully connected layer. 

Since the office dataset is quite small, we do not learn the full network for office experiments and instead we only optimize for fully connected layers initializing with the weights pre-trained on ImageNet. In all of our experiments, we set the feature dimension as $128$. We use stochastic gradient descent to learn the feature function as well as the similarity metric with AdaGrad\cite{adagrad}. We initialize variables with truncated normals having unit variance and use the learning rate \SI{2.5e-4}  and the batch size $256$. 

\subsection{Evaluation Procedure}
We evaluate all algorithms in fully transductive setup following the new evaluation setup of \cite{gong2012}.  We feed training images and labels of the first domain as the source and training images of the second domain as the target. We further evaluate the accuracy on the target domain labels as the ratio of correctly labeled images to all target images.

\subsection{Results}
Following the fully transductive evaluation, we summarize the results in Table~\ref{tab:res} and Table~\ref{tab:res2}. Table~\ref{tab:res} summarizes the results on the object recognition task using office dataset whereas  Table~\ref{tab:res2} summarizes the digit classification task on MNIST and SVHN.

Table~\ref{tab:res}\&\ref{tab:res2} shows results on object recognition and digit classification tasks exhaustively covering all adaptation scenarios. Our algorithm shows state-of-the-art performance. Moreover, our algorithm significantly outperforms all state-of-the-art methods when there is a large domain difference like MNIST$\leftrightarrow$MNIST-M, MNIST$\leftrightarrow$SVHN, Amazon$\leftrightarrow$Webcam and Amazon$\leftrightarrow$D-SLR. We hypothesize this performance is due to the transductive modeling. State-of-the-art algorithms like \cite{ganin15} are seeking for set of features invariant to the domains whereas we seek for an explicit similarity metric explaining both differences and similarities of domains. In other words, instead of seeking for an invariance, we seek for an equivariance.

Table~\ref{tab:res} suggests that accuracy of our algorithm is limited  for D-SLR$\leftrightarrow$Webcam experiments. This is rather expected since the domain difference is very minor between D-SLR and webcam images and the minor domain difference results in saturation of accuracies for all algorithms. Moreover, since we use nearest neighbor classifier, our algorithm needs a large-dataset to be successful. Both webcam and D-SLR datasets are rather small (300 to 700 examples) which limits the accuracy of nearest neighbor algorithm as well.

Table~\ref{tab:res2} further suggests our algorithm is the only one which can generalize that well from MNIST to SVHN dataset. We believe this is thanks to the feature learning in the transductive setup. Clearly the features which are learned from MNIST cannot generalize to SVHN since the SVHN has concepts like color and occlusion which are not available in MNIST. Hence, our algorithm learns SVHN specific features by enforcing accurate transduction in the adaptation stage.

Another interesting conclusion is the asymmetric nature of the results. For example, the accuracy of adapting webcam to amazon and adapting Amazon to webcam is significantly different in Table~\ref{tab:res}. The similar behavior exists in MNIST and SVHN domains as well in Table~\ref{tab:res}. This observation validates the importance of an asymmetric modeling.

\subsubsection{Qualitative Analysis}
To further study the learned representations as well as the similarity metric, we perform a series of qualitative analysis in the form of nearest neighbor analyses and tSNE\cite{tsne} plots.

\begin{figure*}[htb]
    \begin{subfigure}[b]{0.25\textwidth}
        \includegraphics[width=\textwidth]{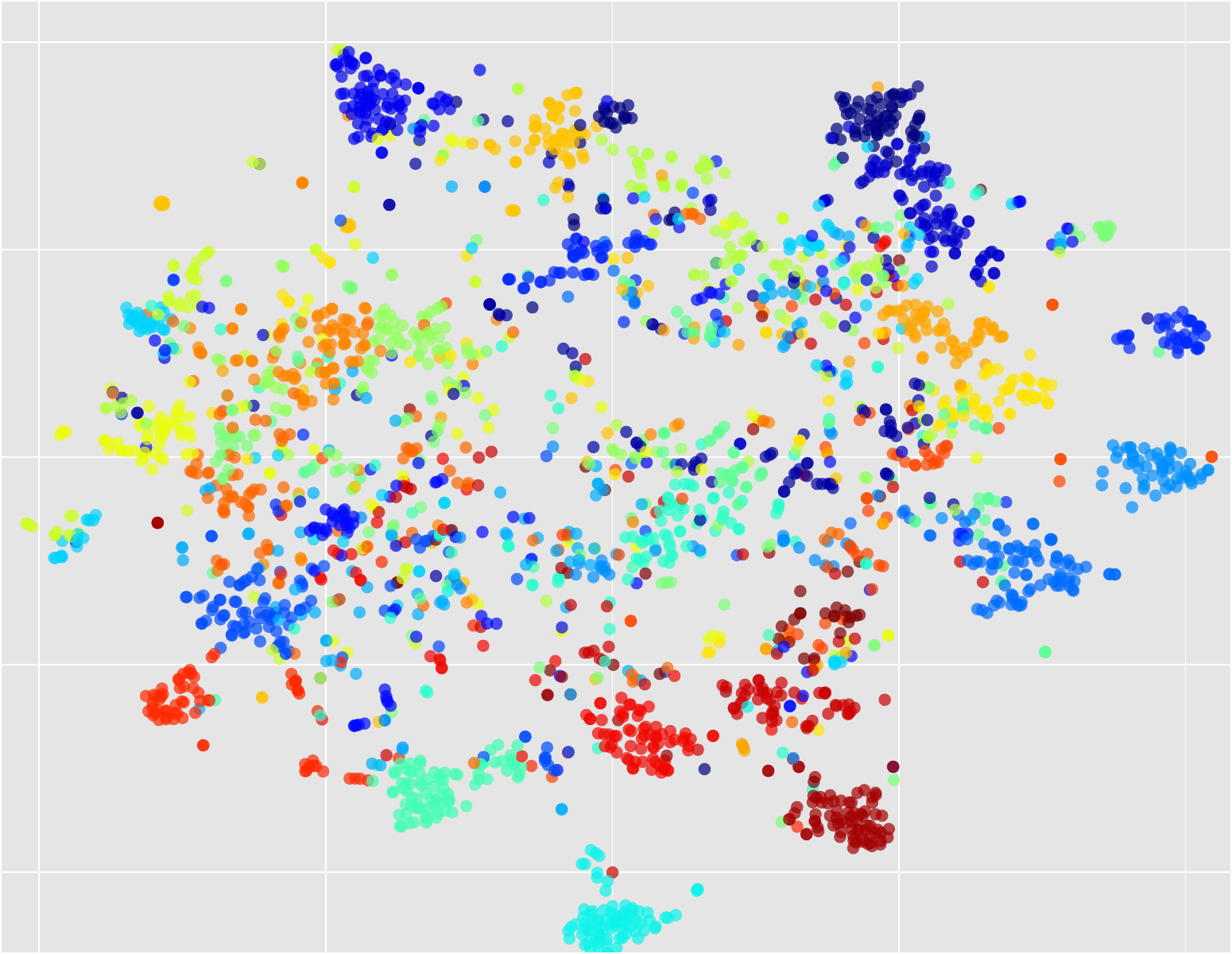}
        \caption{S. w/o Adaptation}
        \label{fig:gull}
    \end{subfigure}~\begin{subfigure}[b]{0.25\textwidth}
        \includegraphics[width=\textwidth]{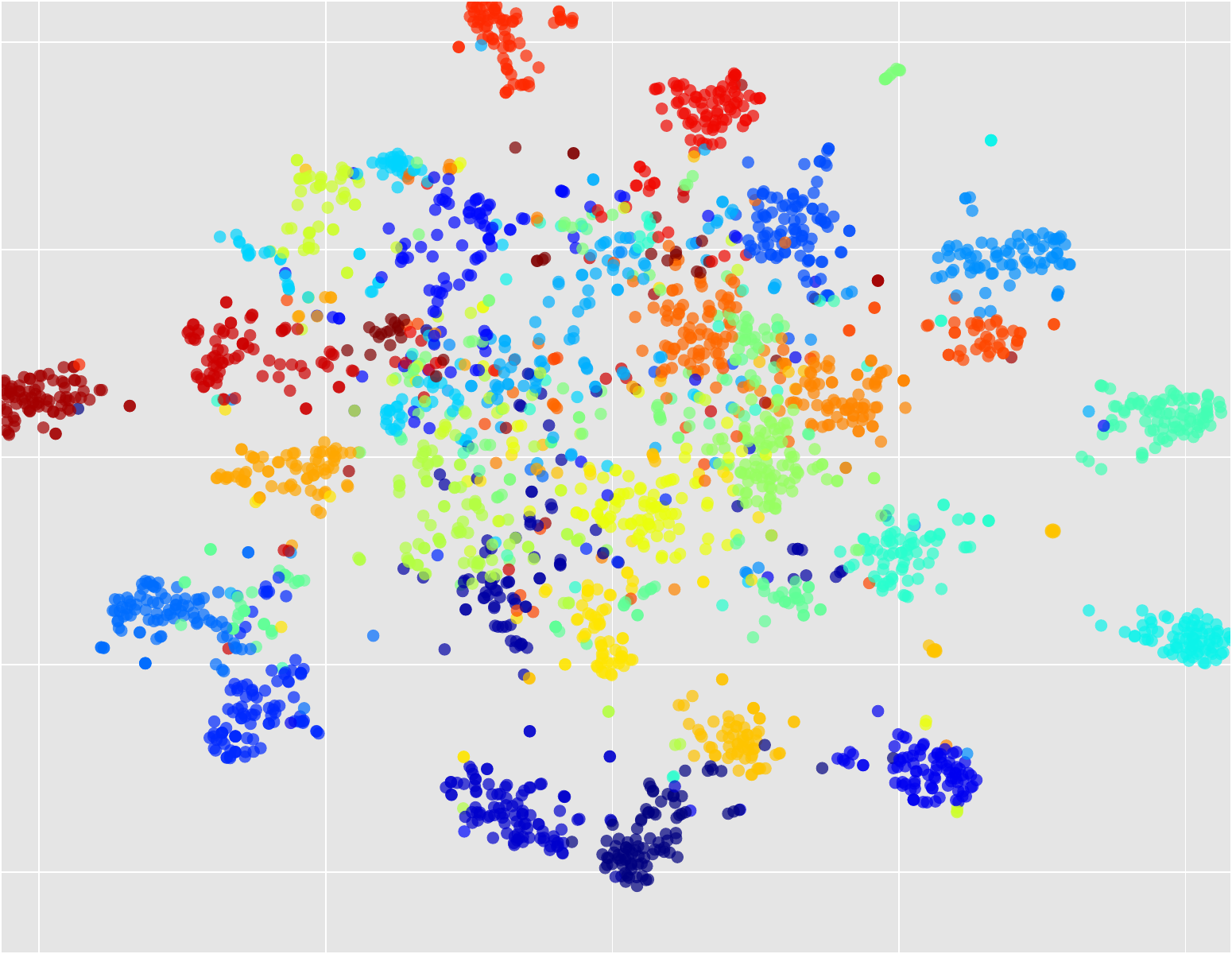}
        \caption{S. with Adaptation}
    \end{subfigure}~\begin{subfigure}[b]{0.25\textwidth}
        \includegraphics[width=\textwidth]{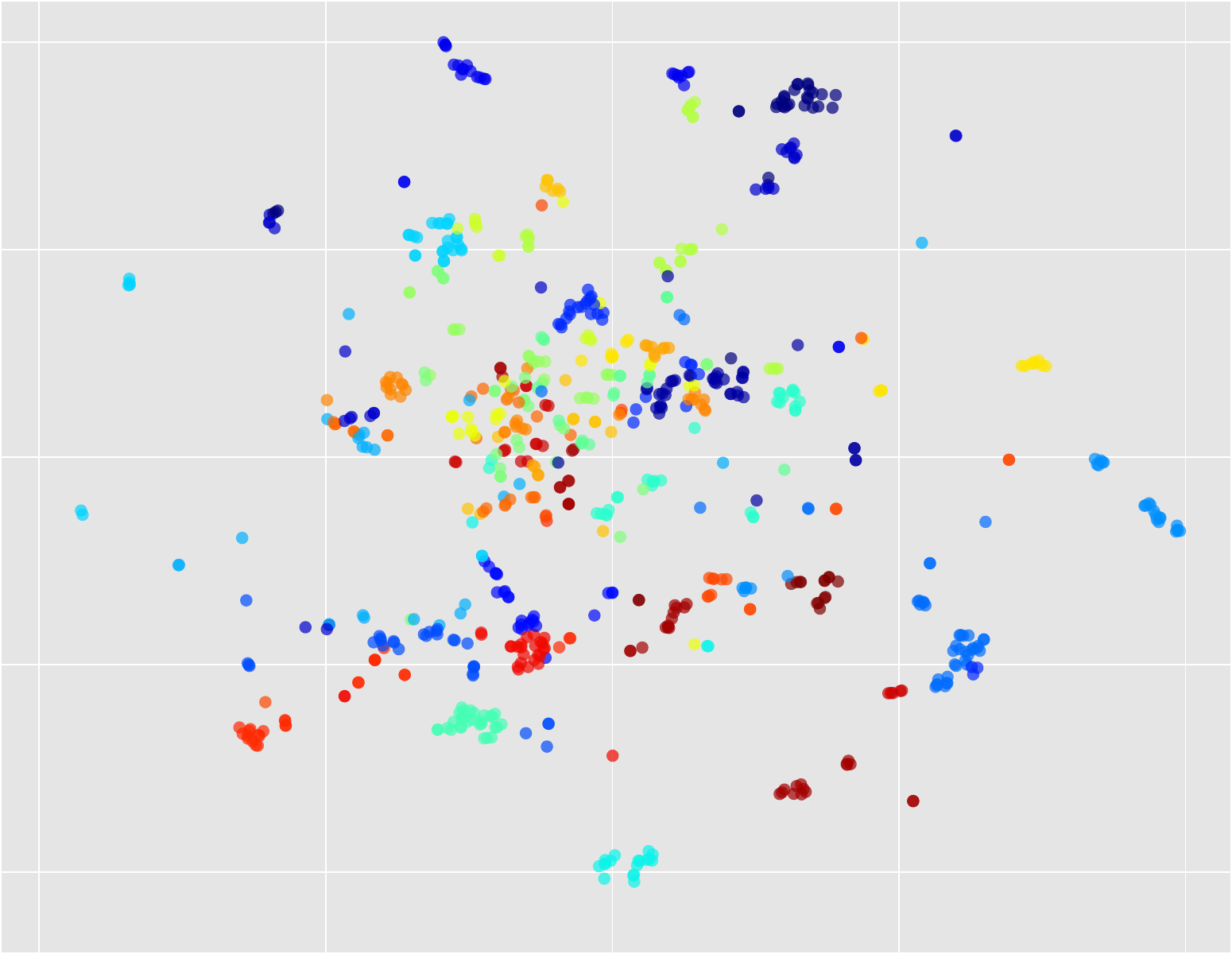}
        \caption{T w/o Adaptation}
        \label{fig:gull}
    \end{subfigure}~\begin{subfigure}[b]{0.25\textwidth}
        \includegraphics[width=\textwidth]{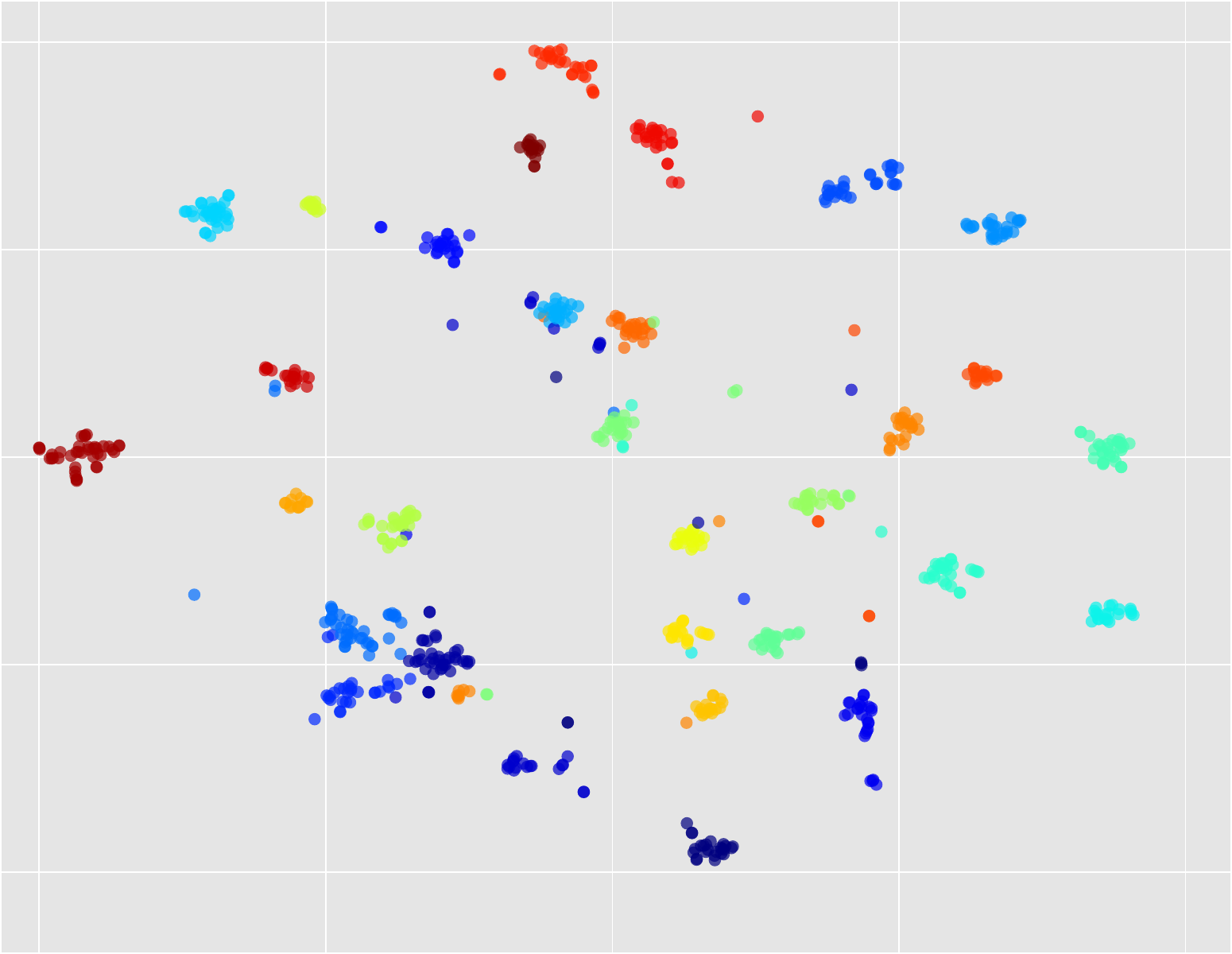}
        \caption{T with Adaptation}
    \end{subfigure}
    \caption{tSNE plots for office dataset Webcam(S)$\rightarrow$Amazon(T). Source features were discriminative and stayed discriminative as expected. On the other hand, target features became quite discriminative after the adaptation.}
    \label{fig:tsne}
        \includegraphics[width=\textwidth]{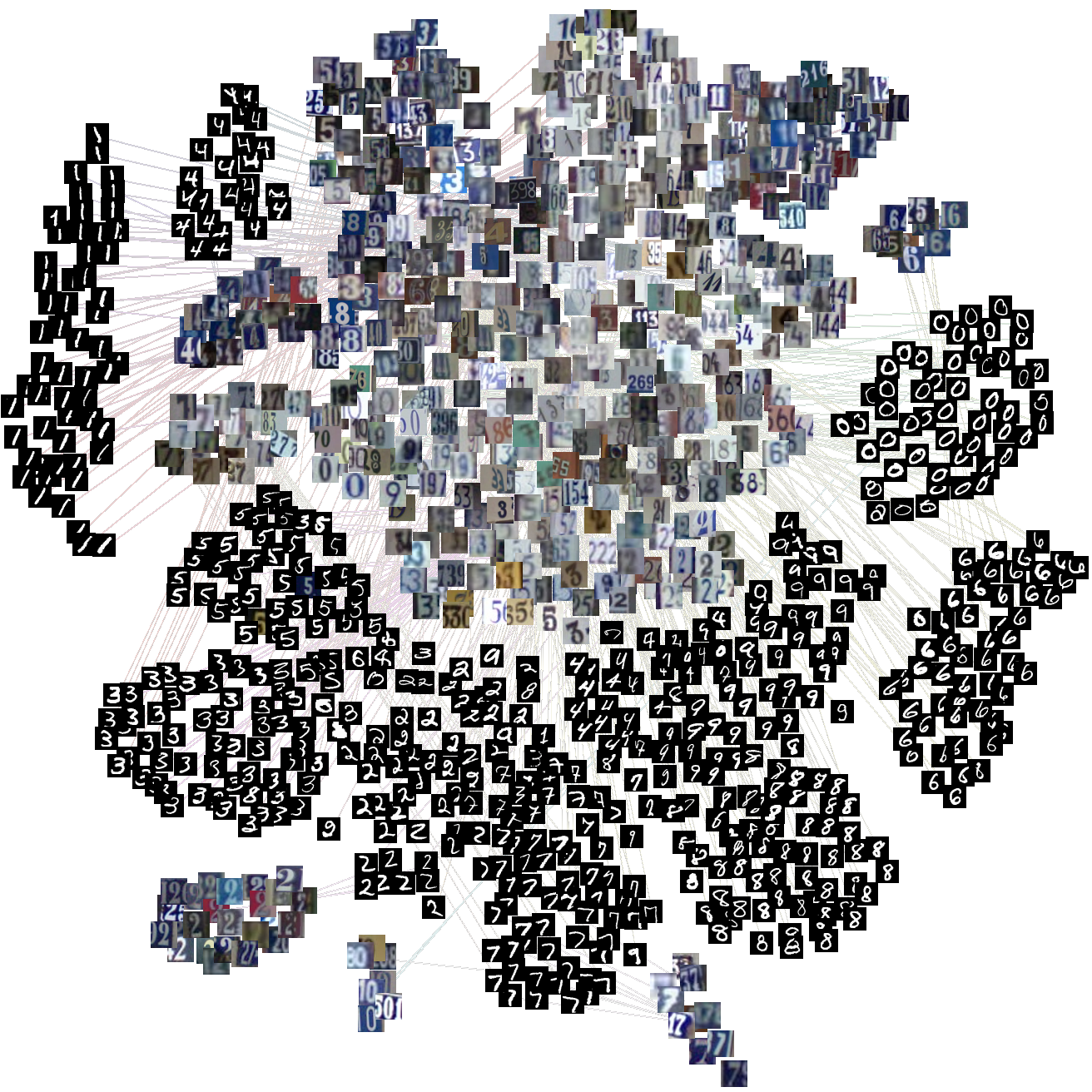}
        \vspace{-5mm}
\caption{tSNE plot for SVHN$\rightarrow$MNIST experiment. Please note that the discriminative behavior only emerges in the unsupervised target instead of the source domain. This explains the motivation behind modeling the problem as transduction. In other words, our algorithm is designed to be accurate and discriminative in the target domain which is the domain we are interested in.  }
\label{fig:tsnedigit}
\end{figure*}

\clearpage
In Figure~\ref{fig:nn}, we visualize example target images from MNIST and their corresponding source images. First of all, both our experimental procedure and qualitative analysis suggest that MNIST and SVHN are the two domains with the largest difference. Hence, we believe MNIST$\leftrightarrow$SVHN is very challenging set-up and despite the huge visual differences, our algorithm results in accurate nearest neighbors.

\begin{figure}[ht]
\begin{small}
\includegraphics[width=\columnwidth]{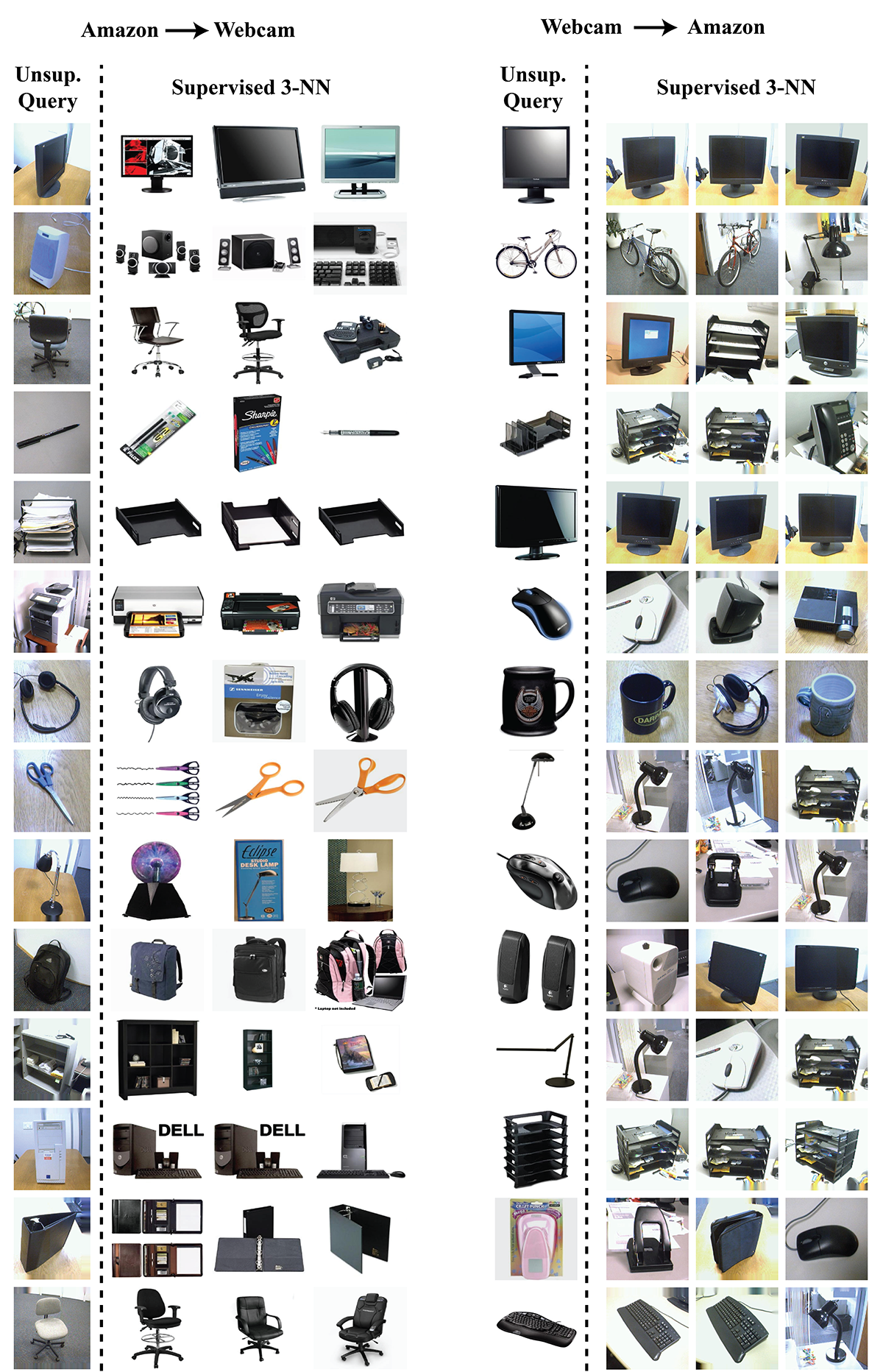}
\caption{Example nearest neighbors for Amazon$\leftrightarrow$Webcam experiment. We show an example source image and 3-NN target images. The drop in the accuracy after the nearest neighbors is expected since our loss function only models the nearest one.}
\label{fig:nnoffice}
\end{small}
\vspace{-5mm}
\end{figure}

In Figure~\ref{fig:nnoffice},  we visualize example target images from webcam and their corresponding nearest source images from Amazon. The accuracy of the domain adaptation is also visible in this task. 

The difference between invariance and equivariance is more clear in the tSNE plots of the Office dataset in Figure~\ref{fig:tsne} as well as digit classification task Figure~\ref{fig:tsnedigit}. In Figure~\ref{fig:tsne}, we plot the distribution of features before and after adaptation for source and target while color coding class labels. As Figure~\ref{fig:tsne} suggests, the source domain is well clustered according to the object classes with and without adaptation. Moreover, this is expected since the features are specifically fine-tuned to the source domain before the adaptation starts. However, target domain features have no structure before adaptation. This is also expected since the algorithm did not see any image from the target domain. After the adaptation, target images also get clustered according to the object classes. 

In Figure~\ref{fig:tsnedigit}, we show the digit images of source and target after the adaptation. Clearly, the target is well clustered according to the classes and source is not very well clustered although it has some structure. Since we learn the entire network for digit classification, our networks learn discriminative features in target domain as our loss depends directly on classification scores in target domain. Moreover, discriminative features in target arises because of the transductive modeling. In comparison, state of the art domain invariance based algorithms only try to be invariant to the domains without explicit modeling of discriminativeness on the target domain. Hence, our similarity metric explicitly models the relationship between the domains and results in an equivariant model while enforcing discriminative behavior in the target. We also draw lines between the nearest neighbor images of source and target images to show the accuracy of the metric function. Moreover, the nearest neighbors are quite accurate confirming the quantitative results. 

\subsubsection{Label propagation \& feature learning}
\begin{figure}[ht]
\vspace{-1mm}
\includegraphics[width=\columnwidth]{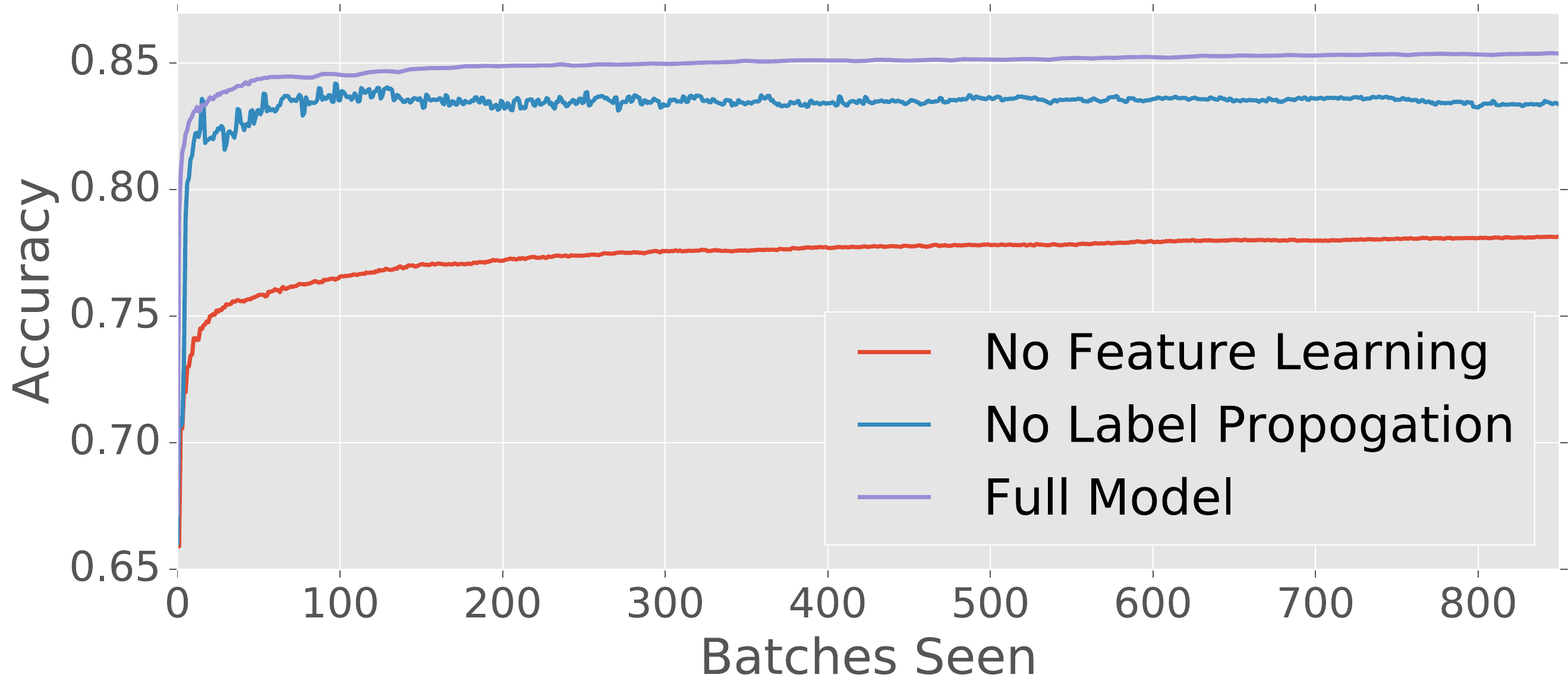}
\vspace{-6mm}
\caption{Accuracy vs number of iterations for our method and its variant without label propagation as well as the variant without feature learning. As the figure suggests the label propagation increases both the stability of the gradients as well as the final accuracy. Moreover, the feature learning also has a significant effect on the accuracy.}
\label{fllprop}
\end{figure}
In order to evaluate the effect of having a robust label propagation and feature learning, we compare our method without the label propagation (noted as \emph{No Label Propogation}) and without feature learning (noted as \emph{No Feature Learning}). We plot the accuracy vs number of iterations in order to evaluate both the effect on learning rate as well as the accuracy. Although we plot the results only for MNIST$\rightarrow$MNIST-M, the other experiments have similar results and not displayed for the sake of clarity.  Results are shown in the Figure~\ref{fllprop}, and it suggests that both feature learning and label propagation is crucial for successful transduction. Another interesting observation is the unstable behavior when we disable label propagation. This is also expected since without label propagation, the labeling stage will have more mis-classifications and they will decrease the accuracy of the metric.

\section{Conclusion} 
We described a transductive approach to the unsupervised domain adaptation problem by defining a joint learning problem on the transductive target label assignment and an asymmetric similarity metric across the domains. We further described a method to learn deep features which are discriminative in the target domain. Experimental results on digit classification using MNIST\cite{mnist} and SVHN\cite{svhn} as well as on object recognition using Office\cite{office} dataset show state of the art performance with a significant margin. 

\bibliographystyle{icml2016}

\end{document}